\documentclass{article}

\usepackage{arxiv}

\usepackage[utf8]{inputenc} 
\usepackage[T1]{fontenc}    
\usepackage{hyperref}       
\usepackage{url}            
\usepackage{booktabs}       
\usepackage{amsfonts}       
\usepackage{nicefrac}       
\usepackage{microtype}      
\usepackage{natbib}
\usepackage{amsmath}
\usepackage{amssymb}
\usepackage{bm}
\usepackage[ruled,vlined]{algorithm2e}

\SetCommentSty{mycommfont}
\usepackage{cleveref}
\usepackage{graphicx}
\usepackage[
  separate-uncertainty = true,
  multi-part-units = repeat
]{siunitx}
\usepackage[section]{placeins}
\usepackage{subcaption}

\title{Reducing Catastrophic Forgetting in Modular Neural Networks by Dynamic Information Balancing}

\author{
  Mohammed Amer\thanks{Corresponding author} \\
  School of Computer Science\\
  University of Nottingham\\
  Semenyih, Malaysia \\
  \texttt{hcxma1@nottingham.edu.my} \\
   \And
  Tom\'as Maul\\
  School of Computer Science\\
  University of Nottingham\\
  Semenyih, Malaysia \\
  \texttt{tomas.maul@nottingham.edu.my} \\
}

\begin{document}
\maketitle

\begin{abstract}

Lifelong learning is a very important step toward realizing robust autonomous artificial agents. Neural networks are the main engine of deep learning, which is the current state-of-the-art technique in formulating adaptive artificial intelligent systems. However, neural networks suffer from catastrophic forgetting when stressed with the challenge of continual learning. We investigate how to exploit modular topology in neural networks in order to dynamically balance the information load between different modules by routing inputs based on the information content in each module so that information interference is minimized. Our dynamic information balancing (DIB) technique adapts a reinforcement learning technique to guide the routing of different inputs based on a reward signal derived from a measure of the information load in each module. Our empirical results show that DIB combined with elastic weight consolidation (EWC) regularization outperforms models with similar capacity and EWC regularization across different task formulations and datasets.

\end{abstract}

\section{Introduction}

Lifelong learning is a key trait that characterises humans and many other animal species in nature and is considered to give a very powerful evolutionary advantage in an ever changing ecosystem which is constantly challenging autonomous learning agents with new survival situations. However, the realization of continual lifelong learning in machine learning systems is still to date an open problem, deep learning systems being no exception. 

Artificial Neural Networks (ANNs), the current workhorse in deep learning systems, are parameterised nonlinear models that are learned by optimizing some objective function through iterative techniques, mostly by stochastic gradient descent (SGD) and its variants. In its most common formulation, lifelong learning in ANNs focuses on learning the model on a sequence of tasks, where each task is defined by its own dataset. Despite the fact that ANNs are able to learn new tasks, their performance on previous tasks tends to degrade. The main underlying cause is the drifting of weights from the optimal point discovered by the learning algorithm in earlier tasks. The term catastrophic forgetting was forged to refer to this phenomenon.

Catastrophic forgetting was described very early in ANN history \citep{McCloskey1989,Ratcliff1990}. The phenomenon is strongly related to the synaptic stability-plasticity dilemma \citep{Abraham2005}, which remains an open problem to date in Neuroscience. According to \citet{Hebb1949}, the learning process in neural circuits is mediated by a change in synaptic strength between neurons through potentiation and depression. This raises the problem of how new experiences are consolidated into synapses without erasing previous experiences, or in other words, how the learning process in the brain strikes the delicate equilibrium point between being plastic enough to allow continual learning, while maintaining sufficient stability so that previous stored information is not damaged. Two main variants of catastrophic forgetting were identified in artificial models: spatial and temporal \citep{Jacobs1991}. Spatial interference happens when a given weight in the network receives conflicting error signals from its outputs. On the other hand, temporal interference is the conflict in the error signals due to different data samples. As far as lifelong learning is concerned, the most impactful factor consists of the temporal interference happening due to different task data distributions.

Five main general approaches were developed for reducing catastrophic forgetting in ANNs, namely, regularization, ensemble, replay, dual-memory and sparse-coding \citep{Kemker2017}. The first approach is based on model regularization, like elastic weight consolidation (EWC) \citep{Kirkpatrick2016} and synaptic intelligence (SI) \citep{Zenke2017}. Given the previous task weights, the main idea is regularizing the weights, based on their importance for the previous task, to stay close to their previous optimal point, while allowing some flexibility to learn the new task. EWC uses Fisher information as a measure of weight importance, while SI uses the path integral of the gradient.




The second approach is ensemble methods \citep{Polikar2001,Dai2007,Fernando}. The motivation behind ensemble methods consists of training multiple classifiers, with a new classifier assigned to each new task, and then integrating their predictions into a final output. The technique in its naive implementation suffers from exploding memory usage as more tasks are learned and inability to transfer learning between tasks, which motivated more research to mitigate these problems.

The third main approach is memory replay \citep{Shin,VandeVen2018,isele2018}. Memory replay is biologically plausible due to the similarity to the suggested hypothesis that the memory consolidating effect of the sleep phase in animals is mediated by the replaying of spike sequences learned during wakefulness \citep{Wei2018}. While regularization focuses on applying direct restriction on the flexibility of weights, memory replay focuses on injecting data samples from previous tasks in order to counteract the tendency of the network to completely shift its learned distribution to the new task, thus forgetting previous experiences. Memory replay techniques differ in the way they acquire previous data samples and how they introduce them into the learning process.


Dual-memory is the fourth approach, which is biologically inspired from memory models in mammalian brains \citep{McClelland1995,Kumaran2016}. The model assumes that two different networks are used for new and old memories, respectively, where newly formed memories are stored in one network and then slowly consolidated into the other network \citep{French1997,Kamra2017}. The consolidation is usually done by simultaneous learning on mixed samples drawn from the new and old memory networks.


Finally come sparse-coding techniques, which are motivated by the idea that interference happens because of overlapping internal representations, hence, introducing carefully-engineered sparsity should in principle reduce this destructive overlapping \citep{Kruschke1991,Coop2013,Murdock1983,Eich1982}. The main limitation of the technique is that sparsity may hinder generalization and model capacity to learn new tasks. 


\Citet{VandeVen2018} identified three main scenarios of continual learning. Task-incremental learning (Task-IL) is a regime of learning sequential tasks, where the model is always provided with task identity. In the second scenario termed domain-incremental learning (Domain-IL), task identity is unknown to the model and the model is required to only solve the task at hand, without explicitly being required to identify the task identity. The last scenario is the class-incremental learning (Class-IL) where the task identity is unknown, however, the model is required to solve the task as well as identify the task identity.

In this paper, we introduce a new technique, that focuses on Task-IL, for reducing catastrophic forgetting by exploiting modular ANNs. The main idea is to distribute the information load between different network modules in order to reduce the ongoing catastrophic forgetting due to interference. In order to do that we need three main components: a way to route information between different modules, a measure to guide the routing based on the information load accumulating in each module and a way of memorizing the path distributions across multiple tasks. The memorization part is necessary for the functioning of the algorithm since the routing is mainly associating an input to a route. The target route will be trained on the routed input and, hence, at inference time similar inputs must be directed to the same route to attain a good performance. For the routing part, we rely on a class of deep models called routing ANNs \citep{McGill2017,Rosenbaum2017,Cai2019}. Routing ANNs are a class of neural network models where input is routed along different paths of the network based on some criteria. In order for the routing to reduce information destruction, we guide the routing process by an approximation of the empirical joint Fisher information using reinforcement learning (RL). The third component, i.e the memory, is realised using a dedicated per task network that is learned in a supervised manner guided by the routing component. Our main contributions in this paper are:

\begin{itemize}
    \item Introducing and investigating the idea of dynamic information balancing (DIB) between different modules in an ANN as a way of alleviating catastrophic forgetting.
    \item Using joint Fisher information as a guiding measure for routing patterns by RL  through modular ANNs in order to reduce information interference.
    \item Approximating empirical joint Fisher information in order to make the routing RL reward tractable.
    \item Introducing a component memory network as a way of preserving path distributions across tasks.
    \item Achieving better generalization results than other models with similar capacity across different task formulations and datasets with diverse distributions.
\end{itemize}

\section{Related Work}

Due to the importance of catastrophic forgetting as a problem affecting our ability to implement continual learning in deep models, many research papers have tried to shed light on the phenomenon and develop techniques for overcoming it. The early work by \citet{Srivastava2013} showed that using local winner-take-all (LWTA) neurons enhances the test error on sequential tasks, suggesting a suppressing effect on catastrophic forgetting. \citet{Goodfellow2013} investigated the effect of dropout and the choice of activation functions on catastrophic forgetting. Their empirical results suggest consistently that training using dropout is a way of reducing catastrophic forgetting across different datasets. On the other hand, different activation functions ranked differently under different conditions, which weakens the argument of a general advantage of the activation function choice on reducing catastrophic forgetting. Dropout is a member of the regularization family of catastrophic forgetting reduction techniques which will be elaborated on more below. On the other hand, activation function selection isn't a mainstream technique for catastrophic forgetting reduction.

Different approaches were suggested for reducing catastrophic forgetting. They can be classified into five main categories: regularization, ensemble, replay, dual-memory and sparse-coding \citep{Kemker2017,Parisi2019}. One of the major contributions in regularization methods consists of the elastic weight consolidation (EWC) technique, introduced by \citet{Kirkpatrick2016}. EWC regularizes the weights by restricting their flexibility so that they don't drift so far from the local minimum discovered in previous tasks. In EWC, the regularization strength of each weight is determined using Fisher information. Synaptic intelligence (SI) \citep{Zenke2017} is another regularization method that uses the path integral of the gradient vector as a strength measure for restricting each weight. Incremental moment matching (IMM) \citep{Lee2017} takes a Bayesian approach and depends on regularizing the posterior distribution's moment of the new task based on the previous task's posterior distribution. \citet{Kaplanis2018} regularize the model through implementing a biologically inspired more complex synapse that takes into account previous modifications applied to the synaptic weight.




Ensemble methods assign an additional model for each new task. Learn++ \citet{Polikar2001,Dai2007} uses algorithms similar to AdaBoost, where a sequential set of classifiers is learned and their predictions are combined using weighted majority voting. PathNet \citep{Fernando} is a form of implicit ensemble method, where genetic algorithms are used to select a pathway through a large network which is trained for the current task. After convergence, the pathway is frozen and another pathway is selected for the next task. A similar implicit technique based on paths is progressive network \citep{Rusu2016}, where a new column is added to a multi-column network for each new task, while the columns of previous tasks are frozen. Each new column is connected to previous columns via lateral connections to promote information reuse. The main limitation for most of the ensemble techniques is the dependence of memory complexity on the number of tasks since a whole model (or a component module) needs to be stored for each task.

Replay methods rely on mixing samples from previous tasks into the learning process to balance the learning process. \citet{Shin} use a generative model, accompanying the main network, that is learned on the data distributions of the previous tasks. The generative model is used to sample inputs from previous tasks, which are mixed with the current task's samples during the training process. \Citet{VandeVen2018} integrate the generative model into the main network by introducing feedback connections that are trained to reconstruct inputs from hidden states, hence, removing the need for a separate generative model. \citet{isele2018} investigate the idea of selecting which experience is more likely to reduce the catastrophic forgetting effect. They investigate four different strategies for selection, namely, surprise, meaning which experience the model finds surprising as measured by the prediction error, reward, which is measured by how strong the reward assigned to the experience is, global distribution matching, which is motivated by capturing the joint strategy for all tasks combined, and coverage maximization, which favors a distribution that covers as much of the input state space as possible.

A cross technique that combines regularization and replay is gradient episodic memory (GEM) proposed by \citet{Lopez-Paz2017}. As in replay, an episodic memory is used for storing samples from previous tasks, but instead of injecting them as a learning input, they are used to regularize the subsequent tasks' learning such that the loss on previous tasks doesn't increase. \cite{Sodhani2019} combines GEM with a network expansion technique called Net2Net \citep{Chen2015} so that when a target performance on a specific task is not achieved, the network is expanded.

Dual-memory approach is biologically inspired and depends on separating the learning of new memories across two different networks, the first is responsible for short-term memories, which are then consolidated into the long-term memory represented by the second network. \citet{French1997} propose using a network composed of two parts, early-processing memory and final-storage memory. The early-processing memory is trained using real data samples and pseudo-samples drawn from the final-storage memory by presenting random inputs. After convergence, the weights of the early-processing memory are transferred to the final-storage memory. The motivation is for the early-processing memory to learn new data samples mixed with data samples drawn from final-storage memory, which already has the old experience consolidated into its data distribution. The approach shares some similarity with replay methods. \citet{Kamra2017} use deep generative models as a way of sampling previous task distributions in a way similar to replay methods. However, their method assigns a new deep generative model to each new task encountered, which are regarded as short-term memories (STMs). After training on multiple tasks, the STMs trained so far are consolidated into a larger long-term memory (LTM) generative network by unsupervised training on the samples from all of the STMs and samples from the LTM itself, which are representatives of old distributions consolidated into the LTM so far.







Sparse-coding methods are based on the assumption that catastrophic forgetting is mainly due to the interference of internal representations, hence, introducing carefully-crafted sparsity will in principle reduce representational overlap. ALCOVE \citep{Kruschke1991} depends on attention gates applied to hidden nodes, such that nodes are activated based on a similarity measure with the input, which is considered as a proxy for task similarity. Another set of algorithms \citep{Murdock1983,Eich1982} store the representations as a superposition between individual states using convolution and correlation as the operators for storage and retrieval, respectively. A fixed expansion layer (FEL) \citep{Coop2013} is a hidden sparse layer initialized in a special way using a mix of fixed excitatory and inhibitory weights. The motivation is to activate different nodes of the layer for different inputs, hence, reducing destructive overlapping.



We rely on dynamic routing for input redirection in DIB. Routing is a way of dynamic module composition conditioned on some criteria, most commonly the input or some representation of it. A routing ANN is a neural network composed of different modules and a routing subnetwork. Given the input, the network is trained to compose a set of modules suitable for handling the input based on the routing subnetwork's decisions, along with the usual weight optimization associated with any ANN. There is neuroscientific evidence that dynamic routing occurs in the primate visual cortex. \citet{Goodale1994} argued that spatial information is processed separately from object identities in the primate visual cortex. 
\citet{McGill2017} use a pyramid of descriptors \citep{Ke2016} as an input to each routing layer and a two way routing subnetwork decides whether to carry on to the next layer, or to stop the signal and produce the final output. The routing subnetwork is trained using different RL algorithms and the training criteria include two penalties for balancing accuracy, i.e more processing, and efficiency, i.e less depth. They regularize only the activated paths to prevent under/over-constraining for frequently/infrequently used paths, respectively. \citet{Rosenbaum2017,Rosenbaum2019} use a global router for doing the routing, which is provided with auxiliary information about the current depth. The training is done using a multi-agent RL (MARL) called weighted policy learner (WPL), which controls the learning rate based on the agent confidence. They experiment with combining two reward signals, a global final reward based on network performance, and a local reward after each action which encourages the agent to minimize the depth of the dynamic path. The recurrent model defined by \citet{Hafner2017} uses routing in a different implicit way. The recurrent model is inspired by the cortico-thalamo-cortico pathway and is composed of different modules each connected to a central routing center. At each time step, different modules read from the routing center using a reading mechanism and all the module outputs are integrated back into the routing center. They argue that the routing is done at the information level in a hierarchical fashion. \citet{Cai2019} propose a neural architecture search (NAS) related approach for routing. During training, different path outputs are combined by a weighted sum using the gumbel-softmax technique \citep{Jang2016,Maddison2016}, while only the top-k paths are selected at inference time.

Related routing mechanisms were introduced by \citet{Sabour2017,Hinton2018}. The routing here is done in the context of capsule networks, which generalizes scalar CNN features to vector representations, termed capsules. The routing is done between different capsules using either routing by agreement \citep{Sabour2017} or expectation-maximization (EM) \citep{Hinton2018}.



In this work, we introduce dynamic information balancing (DIB) as a new method for reducing catastrophic forgetting in modular neural networks (MNNs). DIB combines modular routing ANNs with an approximation of empirical joint Fisher information as a reward signal for routing in order to reduce information destruction. DIB doesn't immediately fall in any of the previous categories mentioned for catastrophic forgetting methods since the core algorithm doesn't use a form of regularization, combine different trained models (ensemble), inject past experience into the training process (replay), use memory consolidation into a global pool (dual-memory) or enforce sparsity (sparse-coding). However, the fact that DIB uses a task specific memory component makes it relatable to explicit ensemble models. However, the memory footprint of DIB is much smaller than ensemble models since only a memory module comprising a small fraction of the model's total number of parameters is stored per task. DIB can also be related to PathNet and implicit ensembling since it exploits different paths through an MNN. However, DIB differs significantly in the routing mechanism and path selection. It is also vaguely related to dual-memory since a form of memory subnetwork is used for each task.

\section{Dynamic Information Balancing}







The aim of dynamic information balancing (DIB) is to reduce destructive information interference by balancing the information content across different modules in a modular neural network (MNN). The main components needed to realise a system like this are:

\begin{itemize}
    \item A modular neural network architecture.
    \item A routing mechanism for routing inputs.
    \item A measure to guide routing such that the information content is balanced.
\end{itemize}




These components need to interact in the following way to achieve information balancing: given an input, the routing mechanism will decide which modules have the least information load, using the information measure of each module, and hence, it will route the input through these modules. After the modules' weights are updated, the information measure of the different modules is updated to reflect the new information load.

\begin{figure}[h]
    \centering
    \includegraphics[width=\textwidth]{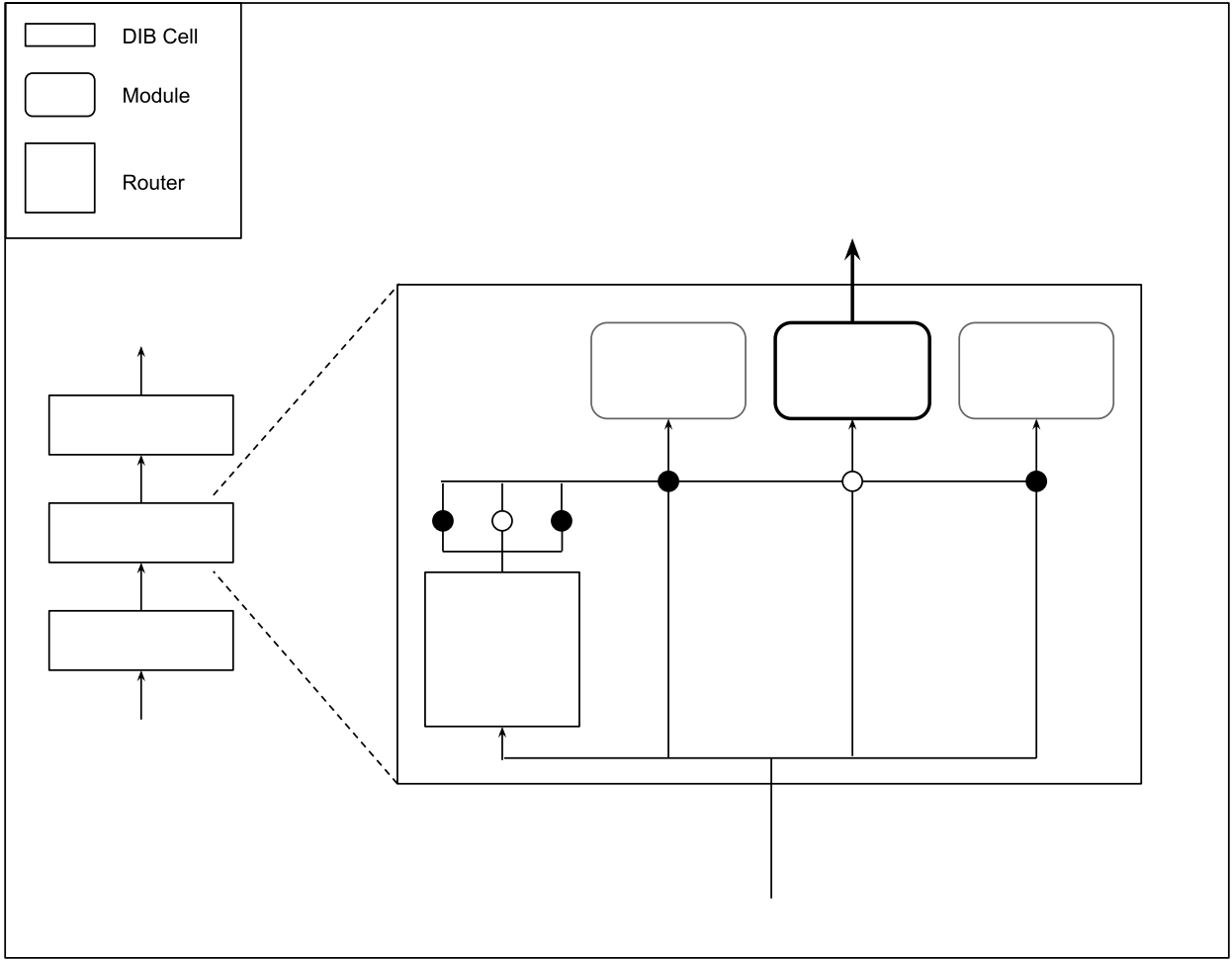}
    \caption{DIB cell architecture (MemNet is omitted). \textbf{White circle:} active connection. \textbf{Black circle:} inactive connection. The router activates one module at a time. At inference time, decisions are made by a MemNet receiving raw input.}
    \label{fig:dib}
\end{figure}

We use a modular architecture that is composed of sequentially stacked DIB cells \cref{fig:dib}. Each DIB cell has three main components. First of all, the set of modules that will be used as a learning substrate. The second component is the router, which is a subnetwork that is used for routing the input to the different modules based on the information measure. The third component is the memory network. The memory network (MemNet) is a subnetwork that is trained to shadow the router by supervised learning using the router's output as a target signal. At inference time, the router is discarded and MemNet is used instead to route inputs.

The need for routing by RL arises from the infeasible combinatorial complexity of searching the space of every possible path and calculating some information measure for each one of these paths. Instead, we rely on a router leaned by RL, which is guided by a reward derived from an information measure.




MemNet is essential for the functioning of the system in the continual learning paradigm. In continual learning, given a new task, we initialize a new set of MemNets (one per cell), that are trained throughout the task by shadowing router decisions. At inference time, the routers are discarded and the MemNets of the task at hand are loaded and used for making routing decisions. The need for MemNet arises from catastrophic interference occurring in the router itself. During task training, the router makes reasonable decisions about input routing based on the modules' information load, but as the input distribution (from the previous DIB cell) shifts between tasks, the router forgets about its past decisions regarding previous distributions. MemNet shadows the router decisions at each task so that any task input can be routed correctly at inference time.

Despite the fact that the MemNet approach has a partial similarity to ensemble methods, its memory requirement is different. The decoupling of the router's and MemNet's architecture allows for using a network that is smaller than the router and much smaller than the total model size, which considerably reduces memory complexity.



\begin{figure}[h]
    \centering
    \includegraphics[width=\textwidth]{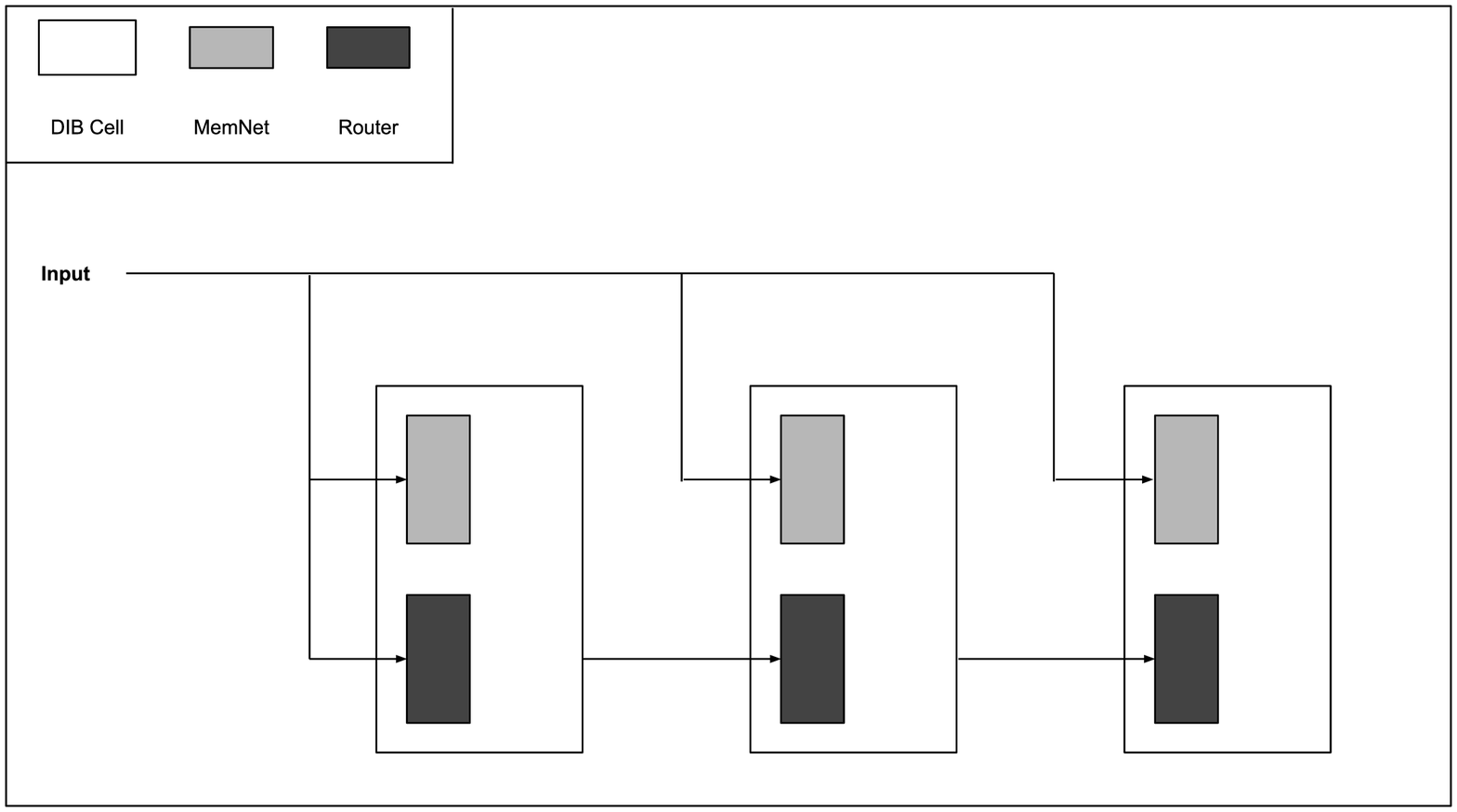}
    \caption{DIB cells wiring diagram (modules are omitted). As depicted, routers receive the output of the previous DIB cell as an input (except in the first cell). In contrast, MemNets always receive the raw input.}
    \label{fig:dib-wiring}
\end{figure}

The router receives the previous DIB cell's output as a conditioning input for making routing decisions. In contrast, MemNet receives the raw input and not the previous DIB cell's output \cref{fig:dib-wiring}. The main reason for this is that the previous DIB cell's output will change upon training on a sequence of tasks. While this change may carry very useful information for the router to make correct decisions, it is confusing to MemNet. MemNet's main task is associating an input pattern with the routing decision at inference time, and, hence, it will not function properly when its inputs get changed.


The router is trained using deep Q-learning (DQN) \citep{Mnih2013}. Given a nonnormalized router's output vector $\pmb{o}_r \in \mathbb{R}^{|\mathbb{A}|}$ where $|\mathbb{A}|$ is the number of possible actions in the action set $\mathbb{A}$ (where each action corresponds to choosing the module with the corresponding index), the router's loss is calculated as:

\begin{equation}
    L_{router} = (r - o_r^{(a)})^2
\end{equation}

where $a$ is the chosen action, $r$ is the reward gained by taking the action $a$ and $o_r^{(a)}$ is the logit of the router's output for the same action. The action is chosen according to an $\epsilon$-greedy policy, where $\epsilon \in [0,1]$ and a random action is chosen with probability $\epsilon$ and the router's optimal action is chosen with probability $1-\epsilon$.








The essence of the DIB is guiding the routing process to reduce information interference. In the assumed DQN RL routing, this translates to using a reward signal that reflects different information loads in each possible action or path. Intuitively, information load of a given module refers to how much information is packed into the specific values of the modules' parameters. Fisher information is a measure of how much information a parameter holds about the distribution which it is modeling. Hence, we use empirical Fisher information with some modifications as a proxy for information load. Given a dataset $\mathbb{S} = \{(\pmb{x}_1,y_1), (\pmb{x}_2,y_2), ..., (\pmb{x}_{|\mathbb{S}|},y_{|\mathbb{S}|})\}$ of size $|\mathbb{S}|$, the usual empirical Fisher information is calculated as:

\begin{equation}
    F_{ii} = \frac{1}{|S|} \sum_{(\pmb{x},y) \in \mathbb{S}} \left(\frac{\partial \log p(Y = y|\pmb{x}; \pmb{\theta})}{\partial \theta_i}\right)^2
\end{equation}

where $\pmb{\theta}$ is the set of model weights and $i$ is the index of the corresponding weight and the fact that we are parameterising by $ii$ reflects that we are using only the diagonal of the Fisher matrix \citep{Kirkpatrick2016}. Due to the nonlinearity applied to the gradient, expressing empirical Fisher as a matrix-matrix operation and, hence, accelerating its calculation through GPUs is infeasible. For regularization purposes like EWC \citep{Kirkpatrick2016}, this is not a serious problem since the Fisher diagonal is only calculated in-between tasks. However, for our purpose of continuously guiding the routing process, we need a continuous efficient way of calculating the empirical Fisher diagonal, otherwise, the router's estimation of information loads will quickly become inaccurate and the routing decisions will start to cause information destruction.

We have done two approximations to allow for continuously using the Fisher information as a routing signal. The first one is approximating the empirical Fisher information by the joint empirical Fisher information over a minibatch. The term joint Fisher information refers to the fact that Fisher information is calculated for the joint probability distribution of multiple samples, instead of the probability of a single sample. Given a minibatch set $\mathbb{B} = \{(\pmb{x}_1,y_1), (\pmb{x}_2,y_2), ..., (\pmb{x}_{|\mathbb{B}|},y_{|\mathbb{B}|})\}$, we calculate the joint Fisher information as:


\begin{equation}
    J_{ii}^{(b)} = \left(\frac{\partial \sum_{(\pmb{x},y) \in \mathbb{B}} \log p(Y = y|\pmb{x}; \pmb{\theta})}{\partial \theta_i}\right)^2
\end{equation}

where $b$ is the batch index. Note that the sum of log probability in the numerator corresponds to the log of the joint probability of the samples in the minibatch under the assumptions of i.i.d. For our purpose, such an approximation is justified by the fact that our routing aims mainly at balancing the information content of the current probability distribution, which can be approximated by the joint probability over a random sample of a minibatch. The second approximation is due to the fact that we are routing over different paths where each path contains many weights, so, we average across all the parameters in the activated modules,

\begin{equation}
    J_M^{(b)} = \frac{1}{|\pmb{\theta}_M|} \sum_{i = 1}^{|\pmb{\theta}_M|} J_{ii}^{(b)}
\end{equation}

where $M$ is the set of activated modules, $\pmb{\theta}_M$ is the set of parameters in the activated modules, $|\pmb{\theta}_M|$ is the number of these parameters and the sum rolls over $i$, which is the index of a parameter in $\pmb{\theta}_M$. 










Since the joint Fisher information we calculated so far is intuitively a measure of the information load in each path, then, the reward should be the negative in this quantity in order to encourage the router to avoid paths with information congestion,

\begin{equation} \label{eq:reward}
    r = -1 \times \lambda \times J_M^{(b)}
\end{equation}

where $\lambda$ is a weighting hyperparameter that controls the strength of the reward. Since this reward is calculated in a minibatch setting, it is actually the reward that will be used for every routing decision in the minibatch. Hence, given the set of routers' actions for minibatch inputs $\mathbb{A}^{(b)}$, the minibatch total routing loss for the given router is calculated as,

\begin{equation}
    L_{router}^{(b)} = \sum_{a \in \mathbb{A}^{(b)}} (r - o_r^{(a)})^2
\end{equation}



\Cref{alg:dib} details how the overall algorithm works. In summary, we loop over the given tasks' datasets. For each epoch in a task's training loop, we calculate the DIB model's outputs on a given minibatch, which includes the classification outputs and router's and MemNet's  outputs for each DIB cell. We calculate the classification loss from the classification outputs and the targets. We calculate the memory loss from the MemNets' outputs and the corresponding routers' outputs. Then, based on the activated modules decided by the routers' decisions, we calculate the joint Fisher information using the classification outputs and the targets. From the minibatch's joint Fisher information, we calculate the reward, which is then used to reward the routers' decisions. Finally, we do the backpropagation and update the relevant parameters for each loss. Note that we have three different losses, one is a classification loss related to the classification outputs, a memory loss related to the supervised training of the MemNets and a routing loss related to the RL of the routers. Classification loss should affect only the modules' weights, memory loss should affect only MemNets' weights and routing loss should affect only the routers' weights.

\begin{algorithm}[H]
\SetAlgoLined

\KwIn{A set of datasets corresponding to different tasks $D = {d_1, d_2, ..., d_T}$ where $T$ is the number of tasks. A DIB model having a set of modules, with the associated routers and MemNets. A reward weight hyperparameter $\lambda$. Training epochs $E$.}

\For{$d_i \gets d_1$ \textbf{to} $d_T$} {
    
    \BlankLine
    initialize a new MemNet
    \BlankLine
    
    \For{$e \gets 1$ \textbf{to} $E$}{
        
        \ForEach{$batch \in d_i$}{
            
            $inputs, targets \gets batch$
            \BlankLine
            
            $class\_outputs, routing\_outputs, mem\_outputs \gets  apply\_DIB\_model(inputs)$
            \BlankLine
            
            $L_{class} \gets classification\_loss(class\_outputs, targets)$
            \BlankLine
            
            $L_{mem} \gets mem\_loss(mem\_outputs,routing\_outputs)$
            \BlankLine
            
            $activated\_modules \gets get\_activated\_modules(routing\_outputs)$
            \BlankLine
            
            $J_M \gets joint\_Fisher(activated\_modules, class\_outputs, targets)$
            \BlankLine
            
            $r \gets -1 \times \lambda \times J_M$
            \BlankLine
            
            $L_{rout} \gets routing\_loss(routing\_output,r)$
            \BlankLine
            
            $backprop\_update\_weights(L_{class}, L_{mem}, L_{rout})$
        }
    }
}

\KwOut{Trained DIB model and a MemNet for each task.}

\caption{DIB}
\label{alg:dib}
\end{algorithm}

\section{Experiments}








We apply DIB to three different Task-IL datasets, two common benchmarks for lifelong learning which are PermutedMNIST and SplitMNIST, and a third more complex camera trap dataset called iWildCam2019, which we preprocess and use in a way similar to SplitMNIST. For each dataset we compare the DIB model against an MLP with a similar capacity. The DIB model we use for all of our experiments is made of two stacked DIB cells, one referred to as hidden cell, the other as output cell. The hidden cell has 10 similar modules, each of which is composed of 2 fully-connected (FC) layers, each of which in turn has 445 neurons. The output cell is composed of 10 modules, each of which is an FC layer with dimensionality matching the task output and a Softmax nonlinearity. The router associated with each cell is an MLP with two hidden layers, each with 256 nodes, and an output layer with dimensionality matching the number of modules (i.e 10 nodes) and Softmax nonlinearity. The MemNet in each cell has two hidden layers, each with 128 nodes and an output layer exactly similar to the router's. We use ReLU activations for all of the hidden nodes. We use an $\epsilon$-greedy policy for the DQN training, with $\epsilon$ initialized to $0$ and updated each training step according to the following formula:

\begin{equation}
    \epsilon_t = \epsilon_{min} + (\epsilon_{max} - \epsilon_{min}) \times \exp(-\lambda \times t)
\end{equation}

where $t$ is the step index, $\epsilon_{min} = 0.1$, $\epsilon_{max} = 1.0$ and $\lambda = 0.001$.



The MLP model we use for comparison under different conditions is designed to have the same depth and almost the same total number of parameters as the DIB model. It is composed of 2 hidden layers, each with 2000 nodes with ReLU activations and an output layer with a Softmax nonlinearity. Since the DIB model depends on a task-specific MemNet and to make a fair comparison with MLP, we add a comparison condition that involves a multi-head MLP (MHMLP). MHMLP has the same architecture mentioned above, except that it has a task-specific output layer. On each new task, we initialize a new output layer, which is trained with the rest of the network. The output layer is stored after the given task's training and reloaded at inference time depending on the task identity. 

We use the mean test error at the final task as a measure of the model performance. This is done by evaluating the model, after training on the final task, on all of the tasks, and then taking the mean. In one of the comparison conditions, we train a vanilla MLP on the whole set of tasks simultaneously. Since such a  model is trained on all tasks jointly, it is an estimation of the lowest error attainable on the given tasks. 

We compare the following conditions for all of the three datasets:

\begin{itemize}
    \item MLP: a vanilla MLP with the above mentioned architecture.
    \item MLP+EWC: an MLP with EWC applied.
    \item MHMLP: a vanilla MHMLP with the above mentioned architecture.
    \item MHMLP+EWC: an MHMLP with EWC applied.
    \item DIB: the DIB model described above.
    \item DIB+EWC: the DIB model with EWC applied to the modules (i.e not applied to the router or MemNet)
    \item lower-bound: an estimation of the test error lower bound, as mentioned above.
\end{itemize}




EWC depends on a weight hyperparameter, which defines the strength of regularization. DIB also depends on a similar hyperparameter that scales the reward signal used for the DQN RL \cref{eq:reward}. Hence, any condition that involves EWC or DIB is run using 5 different hyperparameter values: $(1,10,100,1000,10000)$, and when the condition contains both DIB and EWC, the same values will be used for both DIB and EWC. All our test results are based on the average of 3 trials, and in the case of any condition involving DIB or EWC, the best test performance across all the hyperparameter values is reported. Any single task is trained for 20 epochs and the lower bound condition is trained for 200 epochs. Adam optimizer is used for all of the experiments with default hyperparameter values and a training batch size of 128.

One additional comparison condition, called random information routing (RIR), is done for the SplitMNIST dataset for the sake of analysis and shedding more light on the dynamics of DIB. In RIR, we replace the RL router with random uniform routing of the inputs, i.e a given input is assigned to a module sampled randomly from a uniform distribution over the available modules.

To asses the information content of a specific module, we use conditional-entropy (cond-entropy) as a measure. We calculate the cond-entropy of the final model over all samples, but we accumulate the cond-entropy segregated by which module was activated for each pattern, and then we calculate module-mean cond-entropy as the average cond-entropy of that module over the samples which it was activated for.


\begin{table}[h]
    \centering
    \begin{tabular}{|c|c|}
    
        \multicolumn{2}{l}{PermutedMNIST (lower-bound=\SI{1.76 \pm 0.1}{})}\\
        \hline
        Model & Test error(\%)\\
        \hline
        MLP &  \SI{46.56 \pm 0.16}{}\\
        MLP+EWC & \SI{2.96 \pm 0.1}{}\\
        MHMLP & \SI{43.69 \pm 2.8}{}\\
        MHMLP+EWC & \SI{2.71 \pm 0.28}{}\\
        DIB & \SI{44.6 \pm 1.8}{}\\
        \textbf{DIB+EWC} & \textbf{2.32}\SI{\pm 0.06}{}\\
        \hline
        
    \end{tabular}
    \begin{tabular}{|c|c|}
    
        \multicolumn{2}{l}{SplitMNIST (lower-bound=\SI{0.94 \pm 0.1}{})}\\
        \hline
        Model & Test error(\%)\\
        \hline
        MLP &  \SI{ 44.45\pm 0.5}{}\\
        MLP+EWC & \SI{38.68 \pm 0.5}{}\\
        MHMLP & \SI{31.42 \pm 2.5}{}\\
        MHMLP+EWC & \SI{25.23 \pm 6.2}{}\\
        RIR+EWC & \SI{33.39 \pm 1.2}{}\\
        DIB & \SI{22.96 \pm 9.9}{}\\
        \textbf{DIB+EWC} & \textbf{4.32} \SI{\pm 1.3}{}\\
        \hline
        
    \end{tabular}
    
    \begin{tabular}{|c|c|}
    
        \multicolumn{2}{l}{iWildCam2019 (lower-bound=\SI{8.16 \pm 0.23}{})}\\
        \hline
        Model & Test error(\%)\\
        \hline
        MLP &  \SI{38.34 \pm 1.6}{}\\
        MLP+EWC & \SI{34.03 \pm 0.1}{}\\
        MHMLP & \SI{32.45 \pm 1.7}{}\\
        MHMLP+EWC & \SI{25.06 \pm 2.4}{}\\
        DIB & \SI{36.71 \pm 5.3}{}\\
        \textbf{DIB+EWC} & \textbf{22.74}\SI{\pm 2.9}{}\\
        \hline
        
    \end{tabular}
    \caption{Mean test errors across all tasks.}
    \label{tab:test-error}
\end{table}

\begin{figure*}
    \centering
    \begin{subfigure}[b]{0.475\textwidth}
        \centering
        \includegraphics[width=\textwidth]{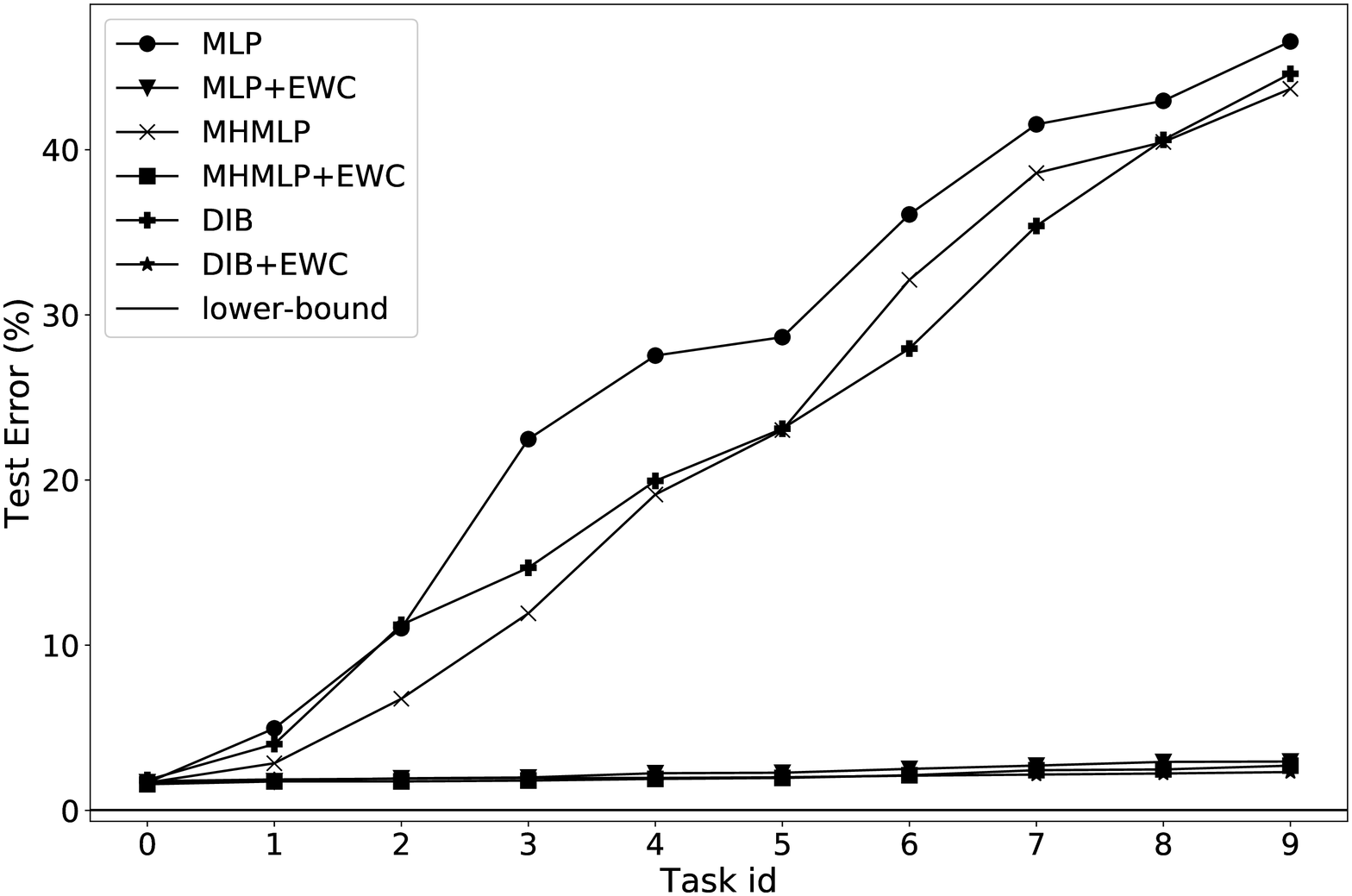}
        \caption[]%
        {{\small PermutedMNIST (all)}}    
        \label{fig:permut-mnist-all}
    \end{subfigure}
    \hfill
    \begin{subfigure}[b]{0.475\textwidth}  
        \centering 
        \includegraphics[width=\textwidth]{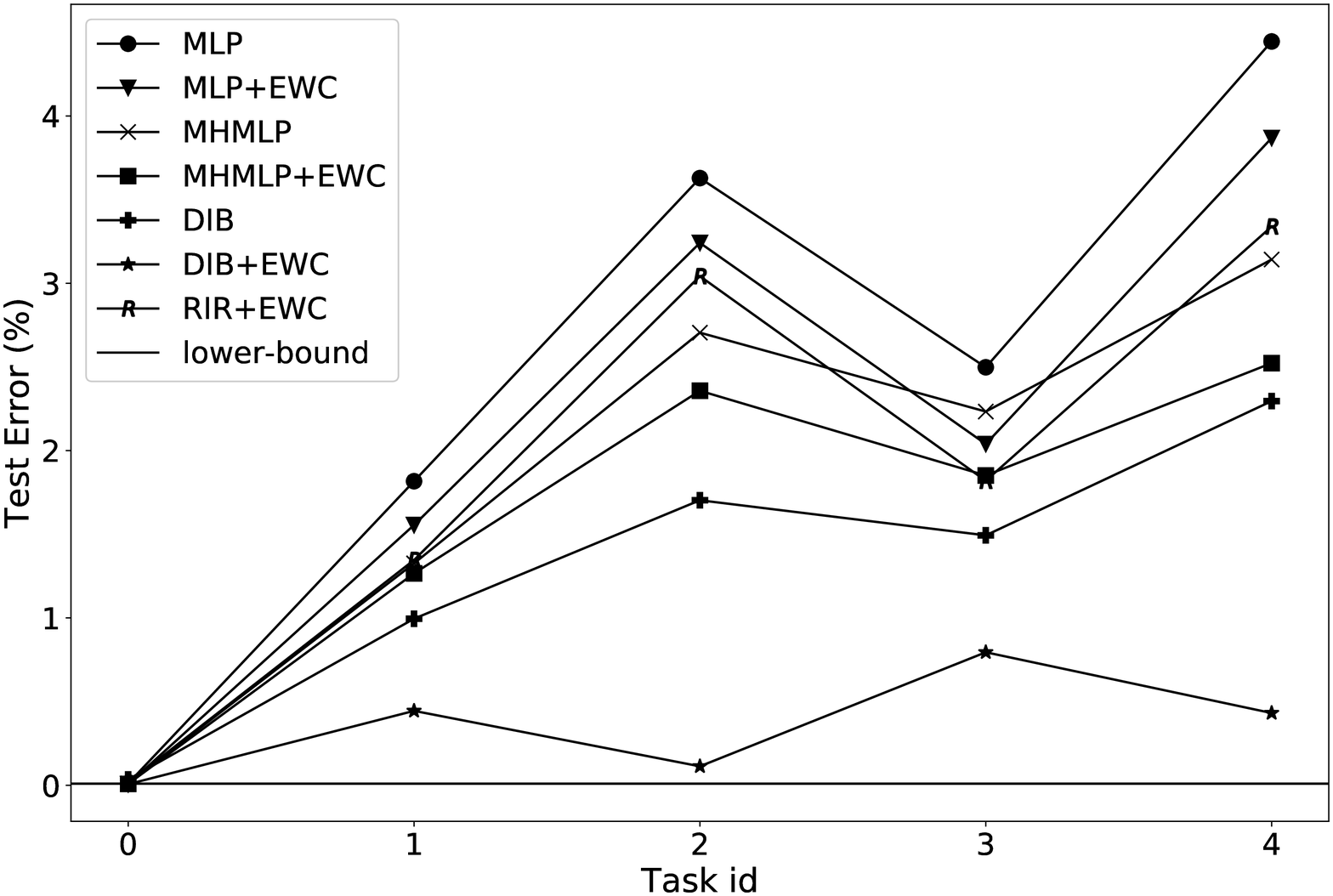}
        \caption[]%
        {{\small SplitMNIST}}    
        \label{fig:split-mnist}
    \end{subfigure}
    \vskip\baselineskip
    \begin{subfigure}[b]{0.475\textwidth}   
        \centering 
        \includegraphics[width=\textwidth]{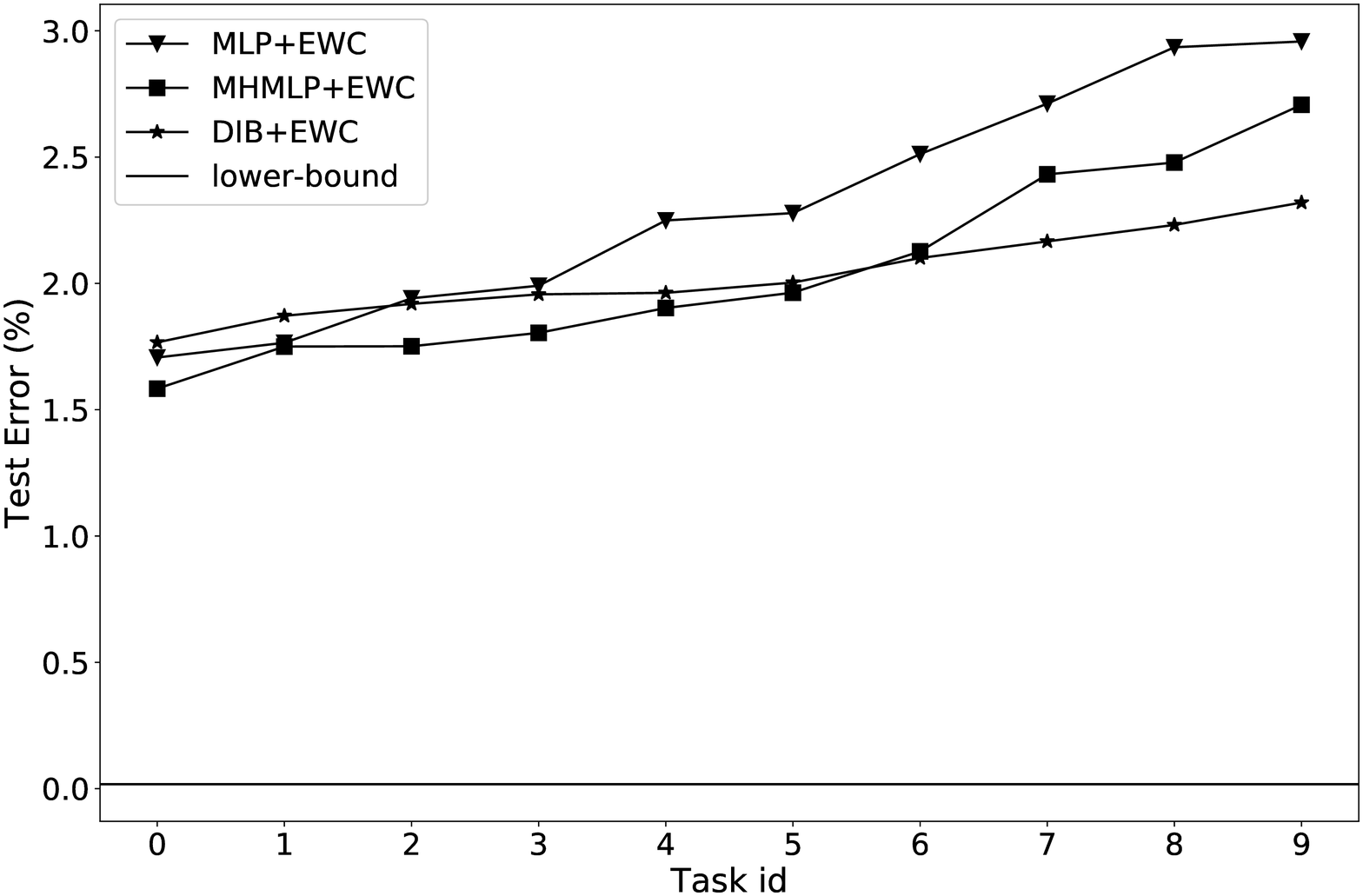}
        \caption[]%
        {{\small PermutedMNIST (zoomed on the best performing)}}    
        \label{fig:permut-mnist-best}
    \end{subfigure}
    \quad
    \begin{subfigure}[b]{0.475\textwidth}   
        \centering 
        \includegraphics[width=\textwidth]{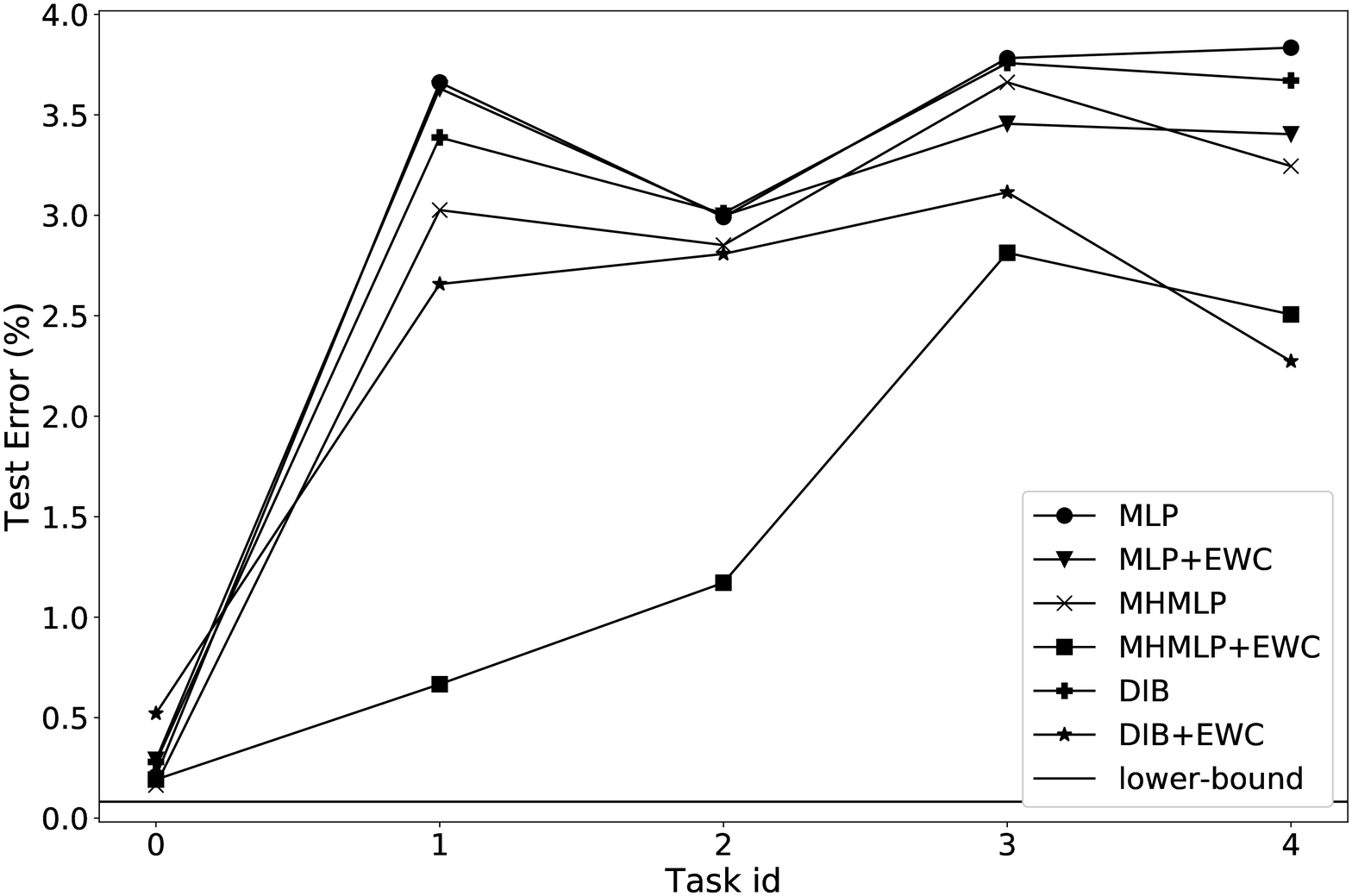}
        \caption[]%
        {{\small iWildCam2019}}    
        \label{fig:iwild}
    \end{subfigure}
    \caption[]
    {\small Test error on previous tasks after training on each task.} 
    \label{fig:test-error}
\end{figure*}

\begin{figure}[h]
    \centering
    \includegraphics[width=0.35\textwidth]{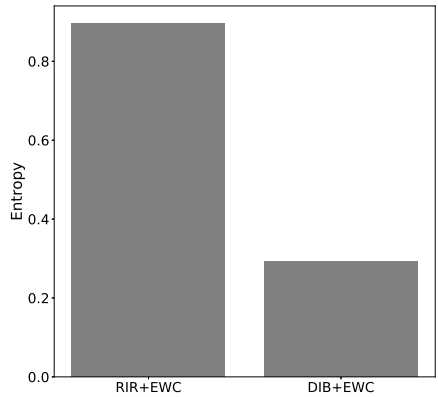}
    \caption{Mean conditional entropy per path.}
    \label{fig:cond-entropy}
\end{figure}






\subsection{Datasets}

\subsubsection{PermutedMNIST}

PermutedMNIST is a classic benchmark for continual learning assessment. The MNIST dataset is used to generate k-number of tasks by shuffling the input pixels by a shuffle order that is applied to all of the inputs in a given task and is different from task to task. The output labels are kept the same. 

We generate 10 tasks, where the first task is just the original dataset, while the remaining 9 tasks are shuffled randomly. The random seed used for shuffling is kept the same for all experiments, to reduce the bias that may be introduced due to the shuffling order. From the training set, we use 90\% for training and 10\% for validation. The test set is used as it is.

\subsubsection{SplitMNIST}

SplitMNIST is based also on the MNIST dataset, however, the tasks are generated by splitting the dataset by combining samples for each two sequential digits together in a disjoint way to generate 5 tasks. We again use 90\% of the task's training data for training and 10\% for validation and we use the partitioned test set as it is.

\subsubsection{iWildCam2019}

We generate tasks from the iWildCam2019 training dataset using a similar way to the SplitMNIST dataset. Before splitting, we have selected 10 of the available classes to generate 5 tasks. We preprocessed the images by gray-scaling and resizing to 88x64. We, then, paired each two classes together and balanced them by discarding the excess samples in the larger class. Because there is no complete overlap between the iWildCam2019 train and test dataset, we used our own test set by dividing each task into 70\% training, 20\% validation and 10\% test. The paired classes are: $[(deer,dog), (squirrel,rodent), (coyote,raccoon), (cat,oppossum), (fox,shunk)]$.  





\subsection{Results}

We benchmark the performance of three main models, namely MLP, MHMLP and DIB, with and without EWC regularization. All the models' performances are enhanced by applying EWC regularization relative to their nonregularized performances across all the datasets \cref{fig:test-error}. DIB+EWC has the best performance across all the datasets \cref{tab:test-error}. In PermutedMNIST \cref{fig:permut-mnist-all}, EWC enhances the different models' performances by a large margin relative to the other two datasets. The performance levels of the EWC conditions have the same following ascending order, from lower to higher performance, across all the datasets: MLP+EWC, MHMLP+EWC then DIB+EWC. The non-EWC conditions don't seem to have a consistent order, however, the condition with the worst performance is MLP across all the datasets \cref{tab:test-error}. The RIR+EWC condition in the SplitMNIST dataset \cref{fig:split-mnist} has lower performance than both the DIB and DIB+EWC. The conditional entropy per path \cref{fig:cond-entropy} is higher in RIR+EWC as compared to DIB+EWC.

\section{Discussion}

The main idea behind DIB is minimizing information interference by routing different patterns through RL rewarded by joint Fisher information. Since the router is guided by an information measure from the current task, the balancing is mainly affecting intra-task interference. This explains the performance gain from EWC regularization, which, despite being a more expensive operation, accounts for inter-task interference. This may also explain the relative performance of DIB+EWC when compared to other EWC conditions. Since other EWC enhanced techniques lack measures for minimizing intra-task interference, they can only adapt to inter-task interference.    

The RIR+EWC condition confirms the effectiveness of RL guided by an information measure. Besides having lower performance than DIB and DIB+EWC, the average entropy per path in RIR+EWC is higher than DIB+EWC. Since EWC is applied to both conditions, i.e RIR+EWC and DIB+EWC, and since EWC regularizes the inter-task interference, then intra-task interference is very likely to contribute significantly to the degraded performance of RIR+EWC.

Another evidence for the contribution of inter-task interference as compared to intra-task interference is the large margin gain in performance in PermutedMNIST compared to the other two datasets. The PermutedMNIST has different input distributions, however, it shares the output distribution. On the other hand, SplitMNIST and iWildCam2019 don't share neither the input nor the output distribution. This shared output distribution between PermutedMNIST tasks means that there is some overlap between tasks, which provides a fertile substrate for inter-task techniques like EWC to significantly reduce the inter-task interference.

One important detail to explain is the choice of the DIB architecture. It may sound more reasonable to diversify the different modules in a DIB cell, instead of choosing homogeneous modules, in order to allow for more differential learning that the routing algorithm can exploit. However, implementing heterogeneous modules isn't straightforward since it can't be readily reduced to a single matrix-matrix operation that can benefit from GPU-acceleration on a minibatch. On the other hand, assuming homogeneity, we could reduce routing different inputs to different paths as a pooled matrix-matrix operation that can be GPU-accelerated using any deep learning library. Despite the fact that we applied the homogeneity assumption to the fully-connected layer, expanding it to other layers like convolutional layers is trivial.





\section{Conclusion}

We have introduced dynamic information balancing (DIB), a method for reducing catastrophic forgetting by dynamically balancing information content across different modules in a modular neural network through routing inputs based on an information theoretic measure. DIB, combined with EWC, achieved better performance than models with similar capacity combined with EWC across different lifelong learning datasets and tasks. We used a computationally cheap approximation of the joint empirical Fisher information as a proxy for information load, which allowed for efficient continual update of the reward needed for guiding the routing by reinforcement learning. MemNet was introduced as a task-specific component that is learned to shadow the router's decisions and take over its routing role at inference time.

We believe there are several potential directions for improving the DIB methodology in the future. The homogeneity assumption, despite the fact that it allowed for an efficient implementation, may be limiting the router's capability of exploiting diversity in the available modules. Finding an efficient generic methodology for the practical realisation of routing efficiently to modules with heterogeneous arbitrary architectures may open the door for a lot of potential enhancements. While relying on task-specific information is a common practice in lifelong learning systems, which is represented by MemNet in our DIB model, finding more task-agnostic ways of reducing catastrophic forgetting is unavoidable for generalizing and extending lifelong learning. We consider extending the information balancing algorithm across the task boundary as a natural generalization for enhancing DIB and reducing the cross-task component of information interference.

\section{Acknowledgement}

This work was partially supported by a grant from Microsoft's AI for Earth program. 

\bibliographystyle{plainnat}
\bibliography{main}

\begin{thebibliography}{44}
\providecommand{\natexlab}[1]{#1}
\providecommand{\url}[1]{\texttt{#1}}
\expandafter\ifx\csname urlstyle\endcsname\relax
  \providecommand{\doi}[1]{doi: #1}\else
  \providecommand{\doi}{doi: \begingroup \urlstyle{rm}\Url}\fi

\bibitem[Abraham and Robins(2005)]{Abraham2005}
Wickliffe~C Abraham and Anthony Robins.
\newblock {Memory retention--the synaptic stability versus plasticity dilemma.}
\newblock \emph{Trends in neurosciences}, 28\penalty0 (2):\penalty0 73--8, feb
  2005.
\newblock ISSN 0166-2236.
\newblock \doi{10.1016/j.tins.2004.12.003}.
\newblock URL \url{http://www.ncbi.nlm.nih.gov/pubmed/15667929}.

\bibitem[Cai et~al.(2019)Cai, Shu, Wang, and Ooi]{Cai2019}
Shaofeng Cai, Yao Shu, Wei Wang, and Beng~Chin Ooi.
\newblock {ISBNet: Instance-aware Selective Branching Network}.
\newblock may 2019.
\newblock URL \url{http://arxiv.org/abs/1905.04849}.

\bibitem[Chen et~al.(2015)Chen, Goodfellow, and Shlens]{Chen2015}
Tianqi Chen, Ian Goodfellow, and Jonathon Shlens.
\newblock {Net2Net: Accelerating Learning via Knowledge Transfer}.
\newblock nov 2015.
\newblock URL \url{http://arxiv.org/abs/1511.05641}.

\bibitem[Coop et~al.(2013)Coop, Mishtal, and Arel]{Coop2013}
Robert Coop, Aaron Mishtal, and Itamar Arel.
\newblock {Ensemble learning in fixed expansion layer networks for mitigating
  catastrophic forgetting}.
\newblock \emph{IEEE Transactions on Neural Networks and Learning Systems},
  24\penalty0 (10):\penalty0 1623--1634, 2013.
\newblock ISSN 2162237X.
\newblock \doi{10.1109/TNNLS.2013.2264952}.

\bibitem[Dai et~al.(2007)Dai, Yang, Xue, and Yu]{Dai2007}
Wenyuan Dai, Qiang Yang, Gui-Rong Xue, and Yong Yu.
\newblock {Boosting for Transfer Learning}.
\newblock Technical report, 2007.
\newblock URL \url{http://www.cs.berkeley.edu/}.

\bibitem[Eich(1982)]{Eich1982}
Janet~M. Eich.
\newblock {A composite holographic associative recall model.}
\newblock \emph{Psychological Review}, 89\penalty0 (6):\penalty0 627--661,
  1982.
\newblock ISSN 0033-295X.
\newblock \doi{10.1037/0033-295X.89.6.627}.
\newblock URL \url{http://content.apa.org/journals/rev/89/6/627}.

\bibitem[Fernando et~al.()Fernando, Banarse, Blundell, Zwols, Ha, Rusu,
  Pritzel, {Wierstra Google DeepMind}, and Brain]{Fernando}
Chrisantha Fernando, Dylan Banarse, Charles Blundell, Yori Zwols, David Ha,
  Andrei~A Rusu, Alexander Pritzel, Daan {Wierstra Google DeepMind}, and Google
  Brain.
\newblock {PathNet: Evolution Channels Gradient Descent in Super Neural
  Networks}.
\newblock Technical report.

\bibitem[French(1997)]{French1997}
Robert~M. French.
\newblock {Pseudo-recurrent Connectionist Networks: An Approach to the
  'Sensitivity-Stability' Dilemma}.
\newblock \emph{Connection Science}, 9\penalty0 (4):\penalty0 353--379, dec
  1997.
\newblock ISSN 09540091.
\newblock \doi{10.1080/095400997116595}.

\bibitem[Goodale et~al.(1994)Goodale, Meenan, B{\"{u}}lthoff, Nicolle, Murphy,
  and Racicot]{Goodale1994}
Melvyn~A. Goodale, John~Paul Meenan, Heinrich~H. B{\"{u}}lthoff, David~A.
  Nicolle, Kelly~J. Murphy, and Carolynn~I. Racicot.
\newblock {Separate neural pathways for the visual analysis of object shape in
  perception and prehension}.
\newblock \emph{Current Biology}, 4\penalty0 (7):\penalty0 604--610, jul 1994.
\newblock ISSN 09609822.
\newblock \doi{10.1016/S0960-9822(00)00132-9}.
\newblock URL
  \url{https://linkinghub.elsevier.com/retrieve/pii/S0960982200001329}.

\bibitem[Goodfellow et~al.(2013)Goodfellow, Mirza, Xiao, Courville, and
  Bengio]{Goodfellow2013}
Ian~J. Goodfellow, Mehdi Mirza, Da~Xiao, Aaron Courville, and Yoshua Bengio.
\newblock {An Empirical Investigation of Catastrophic Forgetting in
  Gradient-Based Neural Networks}.
\newblock dec 2013.
\newblock URL \url{http://arxiv.org/abs/1312.6211}.

\bibitem[Hafner et~al.(2017)Hafner, Irpan, Davidson, and Heess]{Hafner2017}
Danijar Hafner, Alex Irpan, James Davidson, and Nicolas Heess.
\newblock {Learning Hierarchical Information Flow with Recurrent Neural
  Modules}.
\newblock jun 2017.
\newblock URL \url{http://arxiv.org/abs/1706.05744}.

\bibitem[Hebb(1949)]{Hebb1949}
Donald~Olding Hebb.
\newblock \emph{{The organization of behavior}}, volume~65.
\newblock Wiley New York, 1949.

\bibitem[Hinton et~al.(2018)Hinton, Sabour, and Frosst]{Hinton2018}
Geoffrey~E Hinton, Sara Sabour, and Nicholas Frosst.
\newblock {Matrix capsules with EM routing}, feb 2018.
\newblock URL \url{https://openreview.net/forum?id=HJWLfGWRb}.

\bibitem[Isele and Cosgun(2018)]{isele2018}
David Isele and Akansel Cosgun.
\newblock Selective experience replay for lifelong learning, 2018.

\bibitem[Jacobs et~al.(1991)Jacobs, Jordan, and Barto]{Jacobs1991}
Robert~A. Jacobs, Michael~I. Jordan, and Andrew~G. Barto.
\newblock {Task decomposition through competition in a modular connectionist
  architecture: The what and where vision tasks}.
\newblock \emph{Cognitive Science}, 15\penalty0 (2):\penalty0 219--250, apr
  1991.
\newblock ISSN 0364-0213.
\newblock \doi{10.1016/0364-0213(91)80006-Q}.
\newblock URL
  \url{https://www.sciencedirect.com/science/article/pii/036402139180006Q}.

\bibitem[Jang et~al.(2016)Jang, Gu, and Poole]{Jang2016}
Eric Jang, Shixiang Gu, and Ben Poole.
\newblock {Categorical Reparameterization with Gumbel-Softmax}.
\newblock nov 2016.
\newblock URL \url{http://arxiv.org/abs/1611.01144}.

\bibitem[Kamra et~al.(2017)Kamra, Gupta, and Liu]{Kamra2017}
Nitin Kamra, Umang Gupta, and Yan Liu.
\newblock {Deep Generative Dual Memory Network for Continual Learning}.
\newblock oct 2017.
\newblock URL \url{http://arxiv.org/abs/1710.10368}.

\bibitem[Kaplanis et~al.(2018)Kaplanis, Shanahan, and Clopath]{Kaplanis2018}
Christos Kaplanis, Murray Shanahan, and Claudia Clopath.
\newblock {Continual reinforcement learning with complex synapses}.
\newblock In \emph{35th International Conference on Machine Learning, ICML
  2018}, volume~6, pages 3893--3902. International Machine Learning Society
  (IMLS), 2018.
\newblock ISBN 9781510867963.

\bibitem[Ke et~al.(2016)Ke, Maire, and Yu]{Ke2016}
Tsung-Wei Ke, Michael Maire, and Stella~X. Yu.
\newblock {Multigrid Neural Architectures}.
\newblock nov 2016.
\newblock URL \url{http://arxiv.org/abs/1611.07661}.

\bibitem[Kemker et~al.(2017)Kemker, McClure, Abitino, Hayes, and
  Kanan]{Kemker2017}
Ronald Kemker, Marc McClure, Angelina Abitino, Tyler Hayes, and Christopher
  Kanan.
\newblock {Measuring Catastrophic Forgetting in Neural Networks}.
\newblock aug 2017.
\newblock URL \url{http://arxiv.org/abs/1708.02072}.

\bibitem[Kirkpatrick et~al.(2016)Kirkpatrick, Pascanu, Rabinowitz, Veness,
  Desjardins, Rusu, Milan, Quan, Ramalho, Grabska-Barwinska, Hassabis, Clopath,
  Kumaran, and Hadsell]{Kirkpatrick2016}
James Kirkpatrick, Razvan Pascanu, Neil Rabinowitz, Joel Veness, Guillaume
  Desjardins, Andrei~A Rusu, Kieran Milan, John Quan, Tiago Ramalho, Agnieszka
  Grabska-Barwinska, Demis Hassabis, Claudia Clopath, Dharshan Kumaran, and
  Raia Hadsell.
\newblock {Overcoming catastrophic forgetting in neural networks}.
\newblock \emph{arXiv preprint}, 2016.
\newblock ISSN 0027-8424.
\newblock \doi{10.1073/PNAS.1611835114}.

\bibitem[Kruschke(1991)]{Kruschke1991}
John~K. Kruschke.
\newblock Alcove: A connectionist model of human category learning.
\newblock In R.~P. Lippmann, J.~E. Moody, and D.~S. Touretzky, editors,
  \emph{Advances in Neural Information Processing Systems 3}, pages 649--655.
  Morgan-Kaufmann, 1991.
\newblock URL
  \url{http://papers.nips.cc/paper/416-alcove-a-connectionist-model-of-human-category-learning.pdf}.

\bibitem[Kumaran et~al.(2016)Kumaran, Hassabis, and McClelland]{Kumaran2016}
Dharshan Kumaran, Demis Hassabis, and James~L. McClelland.
\newblock {What Learning Systems do Intelligent Agents Need? Complementary
  Learning Systems Theory Updated}.
\newblock \emph{Trends in Cognitive Sciences}, 20\penalty0 (7):\penalty0
  512--534, jul 2016.
\newblock ISSN 13646613.
\newblock \doi{10.1016/j.tics.2016.05.004}.
\newblock URL \url{http://www.ncbi.nlm.nih.gov/pubmed/27315762
  https://linkinghub.elsevier.com/retrieve/pii/S1364661316300432}.

\bibitem[Lee et~al.(2017)Lee, Kim, Jun, Ha, and Zhang]{Lee2017}
Sang-Woo Lee, Jin-Hwa Kim, Jaehyun Jun, Jung-Woo Ha, and Byoung-Tak Zhang.
\newblock {Overcoming Catastrophic Forgetting by Incremental Moment Matching}.
\newblock mar 2017.
\newblock URL \url{http://arxiv.org/abs/1703.08475}.

\bibitem[Lopez-Paz and Ranzato(2017)]{Lopez-Paz2017}
David Lopez-Paz and Marc'Aurelio Ranzato.
\newblock {Gradient Episodic Memory for Continual Learning}.
\newblock jun 2017.
\newblock URL \url{http://arxiv.org/abs/1706.08840}.

\bibitem[Maddison et~al.(2016)Maddison, Mnih, and Teh]{Maddison2016}
Chris~J. Maddison, Andriy Mnih, and Yee~Whye Teh.
\newblock {The Concrete Distribution: A Continuous Relaxation of Discrete
  Random Variables}.
\newblock nov 2016.
\newblock URL \url{http://arxiv.org/abs/1611.00712}.

\bibitem[McClelland et~al.(1995)McClelland, McNaughton, and
  O'Reilly]{McClelland1995}
James~L. McClelland, Bruce~L. McNaughton, and Randall~C. O'Reilly.
\newblock {Why there are complementary learning systems in the hippocampus and
  neocortex: Insights from the successes and failures of connectionist models
  of learning and memory.}
\newblock \emph{Psychological Review}, 102\penalty0 (3):\penalty0 419--457, jul
  1995.
\newblock ISSN 1939-1471.
\newblock \doi{10.1037/0033-295X.102.3.419}.
\newblock URL \url{http://www.ncbi.nlm.nih.gov/pubmed/7624455
  http://doi.apa.org/getdoi.cfm?doi=10.1037/0033-295X.102.3.419}.

\bibitem[McCloskey and Cohen(1989)]{McCloskey1989}
Michael McCloskey and Neal~J. Cohen.
\newblock {Catastrophic Interference in Connectionist Networks: The Sequential
  Learning Problem}.
\newblock \emph{Psychology of Learning and Motivation}, 24:\penalty0 109--165,
  jan 1989.
\newblock ISSN 0079-7421.
\newblock \doi{10.1016/S0079-7421(08)60536-8}.
\newblock URL
  \url{https://www.sciencedirect.com/science/article/pii/S0079742108605368}.

\bibitem[McGill and Perona(2017)]{McGill2017}
Mason McGill and Pietro Perona.
\newblock {Deciding How to Decide: Dynamic Routing in Artificial Neural
  Networks}.
\newblock mar 2017.
\newblock URL \url{http://arxiv.org/abs/1703.06217}.

\bibitem[Mnih et~al.(2013)Mnih, Kavukcuoglu, Silver, Graves, Antonoglou,
  Wierstra, and Riedmiller]{Mnih2013}
Volodymyr Mnih, Koray Kavukcuoglu, David Silver, Alex Graves, Ioannis
  Antonoglou, Daan Wierstra, and Martin Riedmiller.
\newblock {Playing Atari with Deep Reinforcement Learning}.
\newblock dec 2013.
\newblock URL \url{http://arxiv.org/abs/1312.5602}.

\bibitem[Murdock(1983)]{Murdock1983}
Bennet~B Murdock.
\newblock {A Distributed Memory Model for Serial-Order Information}.
\newblock Technical Report~4, 1983.

\bibitem[Parisi et~al.(2019)Parisi, Kemker, Part, Kanan, and
  Wermter]{Parisi2019}
German~I. Parisi, Ronald Kemker, Jose~L. Part, Christopher Kanan, and Stefan
  Wermter.
\newblock {Continual lifelong learning with neural networks: A review}.
\newblock \emph{Neural Networks}, 113:\penalty0 54--71, may 2019.
\newblock ISSN 0893-6080.
\newblock \doi{10.1016/J.NEUNET.2019.01.012}.
\newblock URL
  \url{https://www.sciencedirect.com/science/article/pii/S0893608019300231}.

\bibitem[Polikar et~al.(2001)Polikar, Udpa, Honavar, Member, and
  Udpa]{Polikar2001}
Robi Polikar, Lalita Udpa, Vasant Honavar, Senior Member, and Satish~S Udpa.
\newblock {Learn++: An Incremental Learning Algorithm for Supervised Neural
  Networks mHealth View project Analysis of RNA-protein interactions View
  project Learn++: An Incremental Learning Algorithm for Supervised Neural
  Networks}.
\newblock Technical Report~4, 2001.
\newblock URL \url{https://www.researchgate.net/publication/2489080}.

\bibitem[Ratcliff(1990)]{Ratcliff1990}
R~Ratcliff.
\newblock {Connectionist models of recognition memory: constraints imposed by
  learning and forgetting functions.}
\newblock \emph{Psychological review}, 97\penalty0 (2):\penalty0 285--308, apr
  1990.
\newblock ISSN 0033-295X.
\newblock \doi{10.1037/0033-295x.97.2.285}.
\newblock URL \url{http://www.ncbi.nlm.nih.gov/pubmed/2186426}.

\bibitem[Rosenbaum et~al.(2017)Rosenbaum, Klinger, and Riemer]{Rosenbaum2017}
Clemens Rosenbaum, Tim Klinger, and Matthew Riemer.
\newblock {Routing Networks: Adaptive Selection of Non-linear Functions for
  Multi-Task Learning}.
\newblock nov 2017.
\newblock URL \url{http://arxiv.org/abs/1711.01239}.

\bibitem[Rosenbaum et~al.(2019)Rosenbaum, Cases, Riemer, and
  Klinger]{Rosenbaum2019}
Clemens Rosenbaum, Ignacio Cases, Matthew Riemer, and Tim Klinger.
\newblock {Routing Networks and the Challenges of Modular and Compositional
  Computation}.
\newblock apr 2019.
\newblock URL \url{http://arxiv.org/abs/1904.12774}.

\bibitem[Rusu et~al.(2016)Rusu, Rabinowitz, Desjardins, Soyer, Kirkpatrick,
  Kavukcuoglu, Pascanu, and Hadsell]{Rusu2016}
Andrei~A. Rusu, Neil~C. Rabinowitz, Guillaume Desjardins, Hubert Soyer, James
  Kirkpatrick, Koray Kavukcuoglu, Razvan Pascanu, and Raia Hadsell.
\newblock {Progressive Neural Networks}.
\newblock jun 2016.
\newblock URL \url{http://arxiv.org/abs/1606.04671}.

\bibitem[Sabour et~al.(2017)Sabour, Frosst, and Hinton]{Sabour2017}
Sara Sabour, Nicholas Frosst, and Geoffrey~E Hinton.
\newblock {Dynamic Routing Between Capsules}.
\newblock oct 2017.
\newblock URL \url{http://arxiv.org/abs/1710.09829}.

\bibitem[Shin et~al.()Shin, Lee, Kim, and {Kim Sk T-Brain}]{Shin}
Hanul Shin, Jung~Kwon Lee, Jaehong Kim, and Jiwon {Kim Sk T-Brain}.
\newblock {Continual Learning with Deep Generative Replay}.
\newblock Technical report.

\bibitem[Sodhani et~al.(2019)Sodhani, Chandar, and Bengio]{Sodhani2019}
Shagun Sodhani, Sarath Chandar, and Yoshua Bengio.
\newblock {Toward Training Recurrent Neural Networks for Lifelong Learning}.
\newblock \emph{Neural Computation}, pages 1--34, nov 2019.
\newblock ISSN 0899-7667.
\newblock \doi{10.1162/neco_a_01246}.
\newblock URL
  \url{https://www.mitpressjournals.org/doi/abs/10.1162/neco{\_}a{\_}01246}.

\bibitem[Srivastava et~al.(2013)Srivastava, Masci, Kazerounian, Gomez, and
  Schmidhuber]{Srivastava2013}
Rupesh~Kumar Srivastava, Jonathan Masci, Sohrob Kazerounian, Faustino Gomez,
  and J{\"{u}}rgen Schmidhuber.
\newblock {Compete to Compute}.
\newblock \emph{Nips}, pages 2310--2318, 2013.
\newblock ISSN 10495258.

\bibitem[van~de Ven and Tolias(2018)]{VandeVen2018}
Gido~M. van~de Ven and Andreas~S. Tolias.
\newblock {Generative replay with feedback connections as a general strategy
  for continual learning}.
\newblock sep 2018.
\newblock URL \url{http://arxiv.org/abs/1809.10635}.

\bibitem[Wei et~al.(2018)Wei, Krishnan, Komarov, and Bazhenov]{Wei2018}
Yina Wei, Giri~P. Krishnan, Maxim Komarov, and Maxim Bazhenov.
\newblock {Differential roles of sleep spindles and sleep slow oscillations in
  memory consolidation}.
\newblock \emph{PLoS Computational Biology}, 14\penalty0 (7), jul 2018.
\newblock ISSN 15537358.
\newblock \doi{10.1371/journal.pcbi.1006322}.

\bibitem[Zenke et~al.(2017)Zenke, Poole, and Ganguli]{Zenke2017}
Friedemann Zenke, Ben Poole, and Surya Ganguli.
\newblock {Continual Learning Through Synaptic Intelligence}.
\newblock mar 2017.
\newblock URL \url{http://arxiv.org/abs/1703.04200}.

\end{thebibliography}

\end{document}